# Toward Verifiable Misinformation Detection: A Multi-Tool LLM Agent Framework

Zikun Cui*
Stanford University
Stanford,
USA,
cuizk@alumni.stanford.edu

Tianyi Huang
University of California,
Berkeley,
Berkeley,
USA
tianyihuang@berkeley.edu

Chia-En Chiang
University of California,
Berkeley,
Berkeley
chiaenchiang@berkeley.edu

Cuiqianhe Du
University of California,
Berkeley,
Berkeley,
ducuiqianhe@gmail.com

## Abstract

With the proliferation of Large Language Models (LLMs), the detection of misinformation has become increasingly important and complex. This research proposes an innovative verifiable misinformation detection LLM agent that goes beyond traditional true/false binary judgments. The agent actively verifies claims through dynamic interaction with diverse web sources, assesses information source credibility, synthesizes evidence, and provides a complete verifiable reasoning process. Our designed agent architecture includes three core tools: precise web search tool, source credibility assessment tool and numerical claim verification tool. These tools enable the agent to execute multi-step verification strategies, maintain evidence logs, and form comprehensive assessment conclusions. We evaluate using standard misinformation datasets such as FakeNewsNet, comparing with traditional machine learning models and LLMs. Evaluation metrics include standard classification metrics, quality assessment of reasoning processes, and robustness testing against rewritten content. Experimental results show that our agent outperforms baseline methods in misinformation detection accuracy, reasoning transparency, and resistance to information rewriting, providing a new paradigm for trustworthy AI-assisted fact-checking.

## CCS CONCEPTS

Computing methodologies ~ Intelligent agents

## 1 INTRODUCTION

In the era of information explosion, the rapid spread of misinformation has become a major societal challenge. The widespread adoption of Large Language Models (LLMs) has further exacerbated this issue by making the generation of false information easier and its detection more difficult. For example, during COVID-19 pandemic, the misleading assertion that a "99% survival rate means vaccination is unnecessary" spread extensively on social media platforms. This claim notably disregarded the accuracy of the quoted survival rate and neglected the essential medical context. Similarly, manipulated statistical data on election turnout and climate change have been used to promote false narratives and evade traditional fact-checking methods. Such numerical misinformation exploits people's inherent trust in quantitative data, making it particularly difficult to refute without specialized verification tools. Traditional misinformation detection methods are typically simple binary classification, which fails to provide transparent reasoning or adapt to complex and evolving information environments. To address these limitations, this research proposes an innovative, verifiable misinformation detection LLM agent designed to offer more detailed and credible information assessment mechanisms through dynamic interactions with diverse web resources.

Misinformation detection has long been a significant research direction within the field of natural language processing. Early approaches mainly relied on feature engineering and traditional machine learning algorithms, later evolving to incorporate deep learning and pre-trained language models [1]. However, these methods face several critical challenges: they are typically trained on static datasets, limiting their ability to adapt to emerging forms of misinformation; they lack transparent reasoning processes, making it difficult for users to understand and trust detection results; and they inadequately evaluate the credibility of information sources, often failing to differentiate between evidence of varying quality [2,3,4].

---

* Corresponding author.

With the rise of large LLMs, there's growing interest in using their reasoning ability for fact-checking. However, this method has several limitations. LLMs are prone to hallucination, often lack access to the latest information, and don't have a structured way to collect and evaluate evidence [5]. In particular, their effectiveness drops off when they encounter misinformation that has been carefully reworded or 'laundered' to seem more plausible [6].

The main objective of this research is to develop an intelligent misinformation detection system that can actively seek and evaluate multi-source evidence, provide transparent reasoning processes, assess the credibility of information sources, and remain robust against information rewriting and evolution. Our key innovation is the design of an LLM-based agent architecture capable of executing multi-step verification strategies, maintaining real-time evidence logs, and conducting comprehensive information assessment using various specialized tools. Unlike existing approaches, our agent focuses not only on the final true/false judgment, but also on ensuring the verifiability and transparency of the entire reasoning process.

This research adopts an approach based on the Agentic Reasoning framework [7], which structures the agent's behavior into an iterative cycle of planning, acting, and reflecting. In the context of misinformation detection, this framework empowers the LLM to break down complex verification tasks into sequential subgoals, select and invoke appropriate tools for each subtask, and refine its reasoning based on intermediate outcomes. By combining autonomous decision-making with external tool usage, the Agentic Reasoning framework allows the LLM to go beyond static prediction and perform dynamic, evidence-based fact-checking, significantly enhancing its robustness and interpretability in real-world settings. Our agent architecture comprises three core components: a web search tool, a source credibility assessment tool and a numerical claim verification tool. These tools work collaboratively, enabling the agent to perform complex verification tasks and evaluate the authenticity of information from multiple perspectives.

In our experimental design, we use standard datasets including FakeNewsNet, LIAR, and Covid-19 to compare our approach against state-of-the-art non-agent LLM detectors and standalone LLMs. Our evaluation includes standard classification metrics, assessments of reasoning quality, and robustness tests against rewritten content. Through this comprehensive approach, we aim to introduce a new paradigm in misinformation detection - one that emphasizes evidence verifiability, reasoning transparency, and judgment reliability.

As information authenticity is increasingly critical, this verifiable misinformation detection approach not only contributes to enhancing public media literacy but also offers technical support for fostering a healthier and more trustworthy information ecosystem.

## 2 RELATED WORK

### 2.1 Traditional Machine Learning

Early research on misinformation detection used handcrafted content-based features and classifiers such as logistic regression and decision trees. Pérez-Rosas et al. [8] proposed a method to detect misinformation based on linguistic features. This method analyzes grammatical structure, emotional tone, and linguistic complexity to assess the authenticity of information. While this method yielded promising results on small-scale datasets, it struggled to generalize to diverse misinformation scenarios due to the extensive effort of feature engineering.

### 2.2 Deep Learning Models

With the advancement of deep learning, researchers began to adopt neural network models for misinformation detection. Instead of manually defining features, deep learning models can automatically learn them from data. Convolutional Neural Networks (CNNs) were applied to capture local patterns in text. Recurrent Neural Networks (RNNs), especially LSTM and GRUs, were used to learn sequence information. These deep learning models outperform the earlier feature-based approaches on multiple benchmarks such as FEVER and FakeNewsNet datasets. Deligiannis et al [9] used a graph CNN over news–publisher networks to achieve 94% accuracy, far surpassing feature-based baselines, which reached only 70% accuracy [10].

### 2.3 Pre-trainied Language Models

The success of transformer-based pre-trained language models has further advanced misinformation detection. Model likes BERT are pre-trained on enormous text corpora to learn rich language representation, which can be fine-tuned to classify misinformation. Rohit et al. [11] fine-tuned BERT models on a fake news dataset and achieved ~98.9% accuracy, outperforming earlier CNN and LSTM models. Additionally, some studies have explored multimodal detection methods that integrate textual, visual, and social network information to further enhance detection effectiveness [12].

Despite these advances, deep learning-based methods still exhibit several critical limitations. Deep learning models rely on the internal knowledge of the model, which hinders adaptability to the rapid evolution of misinformation. In addition, deep learning models lack interpretability in detection results, making it difficult for users to understand the judgment basis. These challenges have driven researchers to explore more intelligent, adaptive, and interpretable detection methods. Recent work by Huang et al. [13] proposed combining contrastive learning with implicit stance reasoning to improve misleading text detection on social media, enhancing both semantic differentiation and detection accuracy across domains.

### 2.4 Large Language Models

The emergence of Large Language Models introduced new opportunities for misinformation detection. Instead of task-specific fine-tuning, LLMs can be used in a zero-shot and few-shot prompting manner to perform misinformation detection. This approach leverages LLM's large knowledge base and strong reasoning ability. Early exploration shows both promise and pitfalls. On the one hand, Hu et al. [14] found that GPT 3.5 can identify fake news and generate multi-perspective rationales. Moreover, the LLM-based approach is robust to paraphrases or noise because of its advanced content understanding ability. On the other hand, LLM-based fact-checking methods also face several critical challenges: (1) LLMs cannot access the latest information and relies on parameterized knowledge learned during training; (2) LLMs hallucinate and may generate content that appears reasonable but is factually incorrect; (3) LLMs output can be inconsistent and sensitive to prompting.

### 2.5 Agent Systems in Information Verification Applications

LLM agents combine LLM with an orchestration layer that can plan actions, invoke external tools, and iteratively refine its reasoning. This design allows the model to go beyond its internal parametric knowledge and actively gather fresh evidence—an ability that is particularly valuable for fact-checking tasks.

Recent efforts have adapted LLM agents specifically for misinformation detection and fact-checking. Agent Systems such as WebGPT-FactCheck [15], CheckMate [16], CrossVerifier [17], and Evidence-Hop [18] applied a retrieval–generation loop in which the agent iteratively searches the web, extracts snippets, and produces a veracity judgment. Although these agents outperform zero-shot prompting baselines on FEVER, LIAR, and FakeNewsNet, they still under-explore the evidence space—most stop after one or two queries, leaving many claims unsupported; they employ shallow credibility modeling, relying on coarse domain heuristics and overlooking contextual cues such as author reputation and citation networks; and they offer limited transparency, as intermediate tool calls, evidence selections, and reasoning chains are part of a system output, which hinder auditability.

To address these problems, this research introduces a verifiable misinformation detection agent system based on tool collaboration. Compared to existing methods, our system has several key innovations: (1) integration of multiple specialized tools, including web search, credibility assessment, and numerical claim verification; (2) implementation of comprehensive evidence logging mechanisms to ensure traceability of the reasoning process; and (3) design of multi-step verification strategies to enhance detection accuracy and robustness. These innovations collectively contribute to the development of more trustworthy and transparent misinformation detection systems.

## 3 METHODOLOGY

### 3.1 Agent Workflow Design

To ensure both accuracy and full traceability, our agent employs a structured, three-stage workflow. This process, illustrated in Figure 1, guides the agent sequentially from initial planning to the final synthesis of evidence and conclusion.

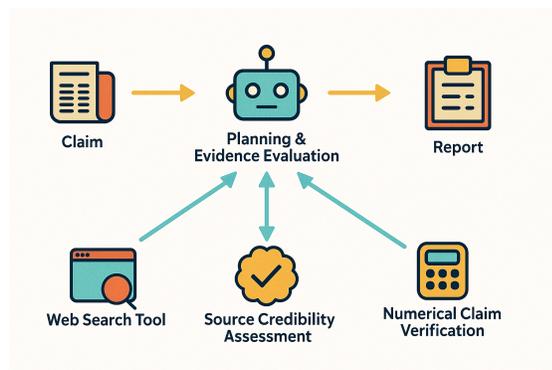

Figure 1. Agent workflow.

#### Stage 1: Planning and Analysis

Upon receiving a claim for verification, the agent begins with comprehensive content analysis. It examines the input to identify core factual claims and numerical assertions that require verification. Based on this analysis, the agent formulates a tailored verification plan that includes:

- The claims that need verification
- The tools to be used for verifying each claim.
- The sequence of tool usage.

This planning stage ensures that the verification process is tailored for each input, allowing the agent to focus on the most critical components of the claim.

**Stage 2: Multi-Tool Execution and Evidence Gathering**

In this stage, the agent follows the plan generated in the last stage to collect evidence using three specialized verification tools, each designed to address a distinct facet of misinformation detection. To enhance the agent's problem-solving ability, our agent is built following the ReAct Framework [19]. During execution, the agent can dynamically update the plan based on its observation from the tool usage result. For example, if the initial web search result doesn't find enough information, it will try to use different search key word to perform another web search attempts. This dynamic reasoning and action taking process allow the agent to solve complex problem like human.

To overcome the limit of the context window of LLM, the system maintains a persistent working memory throughout the verification process, allowing insights from one tool to inform the actions of others. During the final analysis, the agent can selectively retrieve the relevant information stored in the memory to formulate a comprehensive analysis.

**Web Search Tool**: This tool enables the agent to overcome the static limitations of its internal knowledge by gathering real-time information from the internet. After extracting key terms from a claim, the agent uses the tool to conduct web searches and process the resulting web pages to extract the core textual content. This retrieved information is stored in the agent's memory, along with critical metadata like the source URL, publication date, and the original search terms used.

**Source Credibility Assessment Tool**: Because information from different sources carries varying levels of credibility, it must be weighted accordingly in the final analysis. Our Source Credibility Assessment Tool addresses this challenge by evaluating each source using a pre-compiled dataset of credibility ratings. Sources are categorized into three tiers: high-credibility (e.g., established news organizations, academic institutions, government agencies), medium-credibility (e.g., regional publications, specialized industry sources), and low-credibility (e.g., unverified websites, social media posts, anonymous sources). This classification allows the agent to assign an appropriate credibility score to each piece of evidence.

**Numerical Claim Verification Tool**: Recognizing the limitations of LLMs in mathematical reasoning, this specialized tool focuses on verifying numerical claims within the claim. The agent can use this tool to perform calculations to verify mathematical relationships. This addresses a common vulnerability in misinformation where false numerical data is used to support misleading claims.

**Stage 3: Evidence Synthesis and Conclusion Generation**

In the final stage, the agent synthesizes all collected evidence to form a holistic assessment. This process includes:

**Evidence Integration**: The agent evaluates all gathered information, weighing each piece of evidence according to its credibility score and relevance to the original claim. Conflicting evidence is identified and the evidence with higher credibility score is retained.

**Reasoning Chain Construction**: To ensure transparency, the agent constructs a logical reasoning chain that explicitly links the evidence to the conclusion. This chain illuminates how each data point contributes to the overall assessment.

**Conclusion Formulation**: Based on the synthesized evidence, the agent generates a detailed verdict, going beyond a simple true/false binary to include degrees of certainty and identify partially correct claims. This is delivered in a comprehensive report that includes the final results, the step-by-step reasoning process, complete citations for all evidence, and a discussion of any limitations or uncertainties.

## 4 EXPERIMENTS

### 4.1 Experiment Setup

**Datasets:** We selected three public misinformation datasets for evaluation. First, FakeNewsNet serves as a benchmark dataset comprising 23,196 real and 21,417 fake news articles, ideal for binary classification tasks. For a more fine-grained classification challenge, we used LIAR, a more nuanced dataset containing 12,836 political statements labeled across six degrees of truthfulness. Additionally, We used a domain-specific COVID-19 Misinformation dataset containing 5,182 claims related to the pandemic.

To power our source credibility assessment tool, we also curated a source reliability dataset containing credibility ratings for 2,847 distinct sources, including government agencies, news organizations, academic institutions, and social media platforms.

**Agent Implementation:** The agent system is built using the LangGraph framework (v 0.4.9) to orchestrate the multi-step reasoning and tool execution process. The backbone LLMs we use is GPT-4o. The web search Tool is built upon the Tavily web search API, which is tailored for LLM agent use case. The source credibility assessment tool is implemented as a function that cross-references a source's domain against our pre-compiled reliability dataset. Finally, the numerical claim verification tool is a python code interpreter that the agent uses to perform mathematical calculations.

Table 1. Prediction Results Comparison.

| Model Type | Dataset | Accuracy | Precision | Recall | F1 |
|---|---|---|---|---|---|
| SVM | FakeNewsNet | 72.3 | 71.8 | 73.1 | 72.4 |
| Random Forest | FakeNewsNet | 74.6 | 75.2 | 73.9 | 74.5 |
| BERT-base | FakeNewsNet | 81.2 | 82.1 | 80.3 | 81.2 |
| RoBERTa | FakeNewsNet | 83.7 | 84.2 | 83.1 | 83.6 |
| GPT-4o | FakeNewsNet | 85.1 | 85.3 | 84.1 | 84.7 |
| **Our Method** | **FakeNewsNet** | **89.7** | **90.1** | **88.5** | **89.3** |
| SVM | LIAR | 27.8 | 28.3 | 27.2 | 27.7 |
| Random Forest | LIAR | 32.4 | 33.1 | 31.7 | 32.4 |
| BERT-base | LIAR | 52.4 | 58.1 | 56.5 | 57.3 |
| RoBERTa | LIAR | 53.5 | 57.9 | 58.1 | 58.0 |
| GPT-4o | LIAR | 60.6 | 61.2 | 62.7 | 61.9 |
| **Our Method** | **LIAR** | **65.7** | **64.3** | **64.1** | **64.2** |
| SVM | COVID-19 | 60.2 | 60.4 | 61.3 | 60.8 |
| Random Forest | COVID-19 | 64.5 | 63.4 | 64.2 | 63.8 |
| BERT-base | COVID-19 | 78.2 | 77.8 | 76.9 | 77.3 |
| RoBERTa | COVID-19 | 79.5 | 78.8 | 80.4 | 79.6 |
| GPT-4o | COVID-19 | 83.8 | 84.5 | 83.2 | 83.8 |
| **Our Method** | **COVID-19** | **86.2** | **87.1** | **85.4** | **86.2** |

Table 2. Report Quality Evaluation.

| Model Type | Dataset | Relevance | Consistency | Diversity |
|---|---|---|---|---|
| GPT-4o | FakeNewsNet | 0.63 | **0.87** | 0.66 |
| **Our Method** | **FakeNewsNet** | **0.68** | 0.86 | **0.85** |
| GPT-4o | LIAR | 0.72 | 0.85 | 0.56 |
| **Our Method** | **LIAR** | **0.75** | **0.87** | **0.82** |
| GPT-4o | COVID-19 | 0.68 | 0.80 | 0.60 |
| **Our Method** | **COVID-19** | **0.72** | **0.82** | **0.84** |

**Models for Comparison:** Our agent's performance was benchmarked against three categories of models. We first compared it established deep learning models (BERT, RoBERTa). Further comparisons were made with standalone LLM (GPT-4o). Traditional machine learning models (SVM and Random Forest) are also evaluated and serves as a historical baseline. The BERT-based models, SVM ,and Random Forest[Y] are trained on the datasets with a train/eval/test spilt of 70/20/10.

**Evaluation Methods**: The agent output both prediction and the report explaining the reasoning. For the prediction, we evaluate it using accuracy, precision, recall, and F1 score. On the other hand, we use LLM (GPT-4o) as judge to evaluate the quality of the generated reports. For consistency, we use the same prompts to instruct the LLM to evaluate the report and repeat the evaluation threetimes and use the average score as result. The report is evaluated from the following three dimensions.

(1) **Relevance**: assess how relevant the evidence is to the claim. Each evidence is categorized as highly relevant (1.0), slightly relevant (0.5) and irrelevant (0).

(2) **Consistency**: assess the percentage of evidence that is logically consistent the conclusion of the report. Each evidence is categorized as supporting (1.0), irrelevant (0), and contradicting (-1.0).

(3) **Diversity**: assess the diversity of evidence used to supporting the conclusion. Each relevant evidence in the report contribute 0.2 point to the final score and the maximal score is 1.

### 4.2 Experimental Results Analysis

Table 1 demonstrates the performance of our proposed agent system across three datasets. On the FakeNewsNet dataset, our method achieved 89.7% accuracy and an F1 score of 89.3, significantly higher than the machine learning and deep learning method. Compared to standalone LLM, out agent increases F1 score by 4.6%.

Table 3.Tool Ablation Study Results.

| Configuration | FakeNewsNet | | | | LIAR | | | |
| --- | --- | --- | --- | --- | --- | --- | --- | --- |
| | Acc | Pre | Rec | F1 | Acc | Pre | Rec | F1 |
| Full Agent | 89.7 | 90.1 | 88.5 | 89.3 | 65.7 | 64.3 | 64.1 | 64.2 |
| w/o Web Search | 86.1 | 86.6 | 85.2 | 85.9 | 61.5 | 62.0 | 62.9 | 62.4 |
| w/o Credibility Assessment | 87.3 | 88.2 | 86.7 | 87.4 | 63.7 | 63.2 | 63.5 | 63.3 |
| w/o Numerical Verification | 89.2 | 89.5 | 88.3 | 88.9 | 64.9 | 63.9 | 63.8 | 63.8 |
| LLM Only | 85.1 | 85.3 | 84.1 | 84.7 | 60.6 | 61.2 | 62.7 | 61.9 |

On the more challenging dataset (LIAR), the improvement of our method is even more apparent. The LIAR dataset contains six levels of truthfulness, which makes the detection much more difficult. BERT-base deep learning model accuracy is only 53.5%, even GPT-4o can only reach 60.6%. On the hand, our method accuracy reached 65.7%, improving 5.1% compared with GPT-4o and showing that multi-tool collaborative verification method is effective in verifying complex political statements. On the COVID-19 dataset, our method also maintained its lead with 86.2% accuracy, 3.6% improvement compared with GPT-4o.

We also evaluated the report quality of GPT-4o vs. our proposed agent in Table 2. The results show that our agent outperforms GPT-4o on the three datasets in terms of Relevance and Diversity.

For Relevance, our method gets scores of 0.68, 0.75, and 0.72 on three datasets, while GPT-4o gets scores of 0.63, 0.72, and 0.68. The largest improvement comes from Diversity, our agent achieves a significant higher scores of 0.85, 0.82, and 0.84 against 0.66, 0.56, and 0.6 on three datasets. The improvement of diversity score largely benefits from external information from the internet, which our agent can access using the web search tool. Although GPT-4o achieves a higher score on Consistency, our method remains competitive. Overall, these results highlight the advantage of tool-augmented verification in generating more relevant and diverse reports.

### 4.3 Ablation Study

To understand the impact of each individual tool on the system, we perform a comprehensive ablation study by training the following agent variants:

**Full Agent**: Complete agent system containing all three tools

**w/o Web Search**: Agent without the precise web search tool removed

**w/o Credibility Assessment**: Agent without the source credibility assessment tool removed

**w/o Numerical Verification**: Agent without the numerical claim verification tool removed

**Base LLM Only**: Pure LLM baseline without any tools

We evaluate all agents on FakeNewsNet and LIAR datasets with the same metrics.

As shown in Table 3, if we remove the web search tool, we observe the largest performance drop. For FakeNewsNet dataset, the accuracy falls by 3.2% (from 89.7% to 86.1%). In LIAR dataset, the accuracy drops by 4.2%. This shows that external information retrieval is the most important part of our agent, which help mitigates the knowledge limitation of standalone LLM.

Removing the credibility assessment tool leads to a accuracy drop of 2.4% on FakeNewsNet and 2.0% on LIAR, indicating its critical role in misinformation detection. This demonstrates that credibility signals help the agent make more informed judgments, particularly when the textual content alone is ambiguous or misleading.

Removing the numerical verification tool results in a smaller accuracy drop—0.5% on FakeNewsNet and 0.8% on LIAR—yet it remains an essential component. Although numerical claims represent a smaller subset of misinformation, they also demand precise verification to prevent subtle factual errors.

The ablation study highlights the importance of all three tools in achieving optimal system performance. Notably, the combined effect of the full toolset surpasses the additive performance of any pairwise combinations, indicating strong synergistic interactions among the tools. These findings provide evidence for the effectiveness of our multi-tool collaborative architecture.

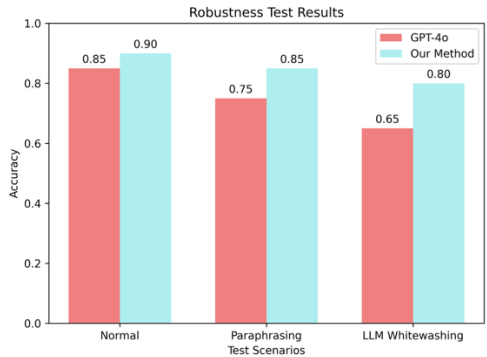

Figure 2. Robustness Test Results.

### 4.4 Robustness Test Results

Figure 2 shows the results of the robustness testing on different levels of information rewriting. When facing simple pharapharsing, our method's accuracy has a minimal decrease of 4.4% (to 85.3%), whereas GPT-4o experienced a substantial drop of 9.4% (to 75.7%). When facing the most challenging "LLM whitewashing" case, our method's accuracy remained strong at 80.2%, dropping only by 9.5%. In contrast, the accuracy of GPT-4o dropped dramatically to 65.7%, representing a 19.4% drop. These results highlight the robustness of our multi-tool collaborative verification mechanisms against malicious information rewriting.

### 4.5 Limitations and Future Directions

While our agent demonstrates strong performance across various misinformation detection datasets , several dataset limitations and biases may affect the generalizability of our findings. The datasets primarily focus on English-language content and U.S.-centric political or health-related misinformation, potentially limiting the applicability of our results to other linguistic, cultural, or topical contexts. Additionally, temporal constraints, such as LIAR's focus on political statements from 2007 to 2016 and rapidly evolving information in the COVID-19 dataset, further restrict the evaluation's scope. Beyond dataset limitations, the credibility assessment tool relies on a predefined set of sources, possibly overlooking emerging credible sources or misclassifying niche but trustworthy providers.

Future directions include addressing these limitations by incorporating more diverse, multilingual, and dynamically updated datasets. Additional efforts should focus on extending the agent framework to handle multimodal misinformation detection, integrating visual and audio analysis to combat increasingly prevalent multimedia misinformation. Improving real-time adaptability and scalability for large-scale deployment, along with developing interactive user interfaces for enhanced transparency and user trust, are also promising areas for future exploration.

### 5 CONCLUSION

In this study, we present a verifiable misinformation-detection agent composed of a LLM and three collaborating tools: web search tool, source-credibility assessment tool, and numerical-claim verification tool. The tool collaboration provides end-to-end, evidence-grounded judgments. By making the verification process explicit as an plan-act-reflect loop and maintaining a persistent evidence log, the agent successfully makes the fact-checking process, which was previously a black box opaque prediction, transparent and auditable reasoning chain.

Experimental results on FakeNewsNet, LIAR, and COVID-19 dataset demonstrate the significant benefits of our multi-tool collaboration. Compared to state-of-the-art LLM baselines, our system achieves an additional 4.6 points of F1 score on FakeNewsNet, 3.7 points on the six-level LIAR task, and 3.6 points on COVID-19 dataset. Quality assesment on the reports generated by our system further show that the reports provide higher levels of relevance and diversity of evidence without sacrificing logical consistency.

Overall, our results demonstrate that careful designed tool collaboration improves both accuracy and trustworthiness of LLM-based fact-checking. The agent framework we propose in this work provides a practical solution for deploying transparent and verifiable AI in practical information environments where credibility and auditability are critical.